\documentclass[runningheads]{llncs}
\usepackage{amsmath,amssymb} 
\usepackage[width=122mm,left=12mm,paperwidth=146mm,height=193mm,top=12mm,paperheight=217mm]{geometry}

\usepackage{graphicx}
\usepackage{color}
\usepackage{times}
\usepackage{epsfig}
\usepackage{graphicx}
\usepackage{amsmath}
\usepackage{amssymb}
\usepackage{mathrsfs}
\usepackage{eurosym}
\usepackage{tabularx}
\usepackage{siunitx}
\usepackage{authblk}

\begin{document}
\pagestyle{headings}
\mainmatter
\def\ECCV18SubNumber{***}  
\title{Learning 3D Human Pose from Structure and Motion}

\titlerunning{ }

\authorrunning{ }

\author{Rishabh Dabral$^1$, Anurag Mundhada$^1$, Uday Kusupati$^1$, Safeer Afaque$^1$, Abhishek Sharma$^2$, Arjun Jain$^1$}

\institute{$^1$Indian Institute of Technology Bombay, $^2$Gobasco AI Labs\\
    {\tt\small \{rdabral, safeer, ajain\}@cse.iitb.ac.in, \{anuragmundhada, kusupatiuday, abhisharayiya\}@gmail.com }}

\maketitle

\begin{abstract}
3D human pose estimation from a single image is a challenging problem, especially for in-the-wild settings due to the lack of 3D annotated data. We propose two anatomically inspired loss functions and use them with the weakly-supervised learning framework of~\cite{Zhou_2017_ICCV} to jointly learn from large-scale in-the-wild 2D and indoor/synthetic 3D data. We also present a simple temporal network that exploits temporal and structural cues present in predicted pose sequences to temporally harmonize the pose estimations. We carefully analyze the proposed contributions through loss surface visualizations and sensitivity analysis to facilitate deeper understanding of their working mechanism. Our complete pipeline improves the state-of-the-art by 11.8\% and 12\% on Human3.6M and MPI-INF-3DHP, respectively, and runs at 30 FPS on a commodity graphics card. 

\end{abstract}
\vspace{-2em}
\section{Introduction}

Accurate 3D human pose estimation from monocular images and videos is the key to unlock several applications in robotics, human computer interaction, surveillance, animation and virtual reality. These applications require \emph{accurate} and \emph{real-time} 3D pose estimation from monocular image or video under challenging variations of clothing, lighting, view-point, self-occlusions, activities, background clutter etc.~\cite{sminchisescu2003estimating,SARAFIANOS20161}. With the advent of recent advances in deep learning, compute hardwares and, most importantly, large-scale \emph{real-world} datasets (ImageNet~\cite{2014arXiv1409.0575R}, MS COCO~\cite{MSCOCO:2014}, CityScapes~\cite{cordts2016cityscapes} etc.), computer vision systems have witnessed dramatic improvements in performance. Human-pose estimation has also benefited from synthetic and real-world datasets such as MS COCO~\cite{MSCOCO:2014}, MPII Pose~\cite{andriluka14cvpr}, Human3.6M~\cite{h36m_pami,IonescuSminchisescu11}, MPI-INF-3DHP~\cite{mono-3dhp2017}, and SURREAL~\cite{Varol_2017_CVPR}. Especially, 2D pose prediction has witnessed tremendous improvement due to large-scale in-the-wild datasets~\cite{MSCOCO:2014,andriluka14cvpr}. However, 3D pose estimation still remains challenging due to severely under-constrained nature of the problem and absence of any real-world 3D annotated dataset.

A large body of prior art either directly regresses for 3D joint coordinates~\cite{li20143d,Li_2015_ICCV,Sun_2017_ICCV} or infers 3D from 2D joint-locations in a two-stage approach~\cite{mono-3dhp2017,Moreno-Noguer_2017_CVPR,Lin_2017_CVPR,zhou2016sparseness,Zhou_2017_ICCV}. These approaches perform well on synthetic 3D benchmark datasets, but lack generalization to the real-world setting due to the lack of 3D annotated in-the-wild datasets. 
To mitigate this issue, some approaches use synthetic datasets~\cite{ChenWLSWTLCC16,Varol_2017_CVPR}, green-screen composition~\cite{mono-3dhp2017,VNect_SIGGRAPH2017}, domain adaptation~\cite{ChenWLSWTLCC16}, transfer learning from intermediate 2D pose estimation tasks~\cite{mono-3dhp2017,li20143d}, and joint learning from 2D and 3D data~\cite{Zhou_2017_ICCV,Sun_2017_ICCV}. Notably, joint learning with 2D and 3D data has shown promising performance in-the-wild owing to large-scale real-world 2D datasets. We seek motivation from the recently published joint learning framework of Zhou et al.~\cite{Zhou_2017_ICCV} and present a novel structure-aware loss function to facilitate training of Deep ConvNet architectures using both 2D and 3D data to accurately predict the 3D pose from a single RGB image. The proposed loss function is applicable to 2D images during training and ensures that the predicted 3D pose does not violate anatomical constraints, namely joint-angle limits and left-right symmetry of the human body. We also present a simple learnable temporal pose model for pose-estimation from videos. The resulting system outperforms the best published system by 12\% on both Human3.6M and MPI-INF-3DHP and runs at 30fps on commodity GPU.

Our proposed structure-aware loss is inspired by anatomical constraints that govern the human body structure and motion. We exploit the fact that certain body-joints cannot bend beyond an angular range; e.g. the knee(elbow) joints cannot bend forward(backward). We also make use of left-right symmetry of human body and penalize unequal corresponding pairs of left-right bone lengths. Lastly, we also use the bone-length ratio priors from~\cite{Zhou_2017_ICCV} that enforces certain pairs of bone-lengths to be constant. It is important to note that the illegal-angle and left-right symmetry constraints are complementary to the bone-length ratio prior, and we show that they perform better too. One of our contributions lies in formulating a loss function to capture joint-angle limits from an inferred 3D pose. We present the visualization of the loss surfaces of the proposed losses to facilitate a deeper understanding of their workings. The three aforementioned structure losses are used to train our \emph{Structure-Aware PoseNet}. Joint-angle limits and left-right symmetry have been used previously in the form of optimization functions~\cite{akhter2015pose,HERDA2005189,bogo2016keep}. To the best of our knowledge we are the first ones to exploit these two constraints, in the form of differentiable and tractable loss functions, to train ConvNets directly. Our structure-aware loss function outperforms the published state-of-the-art in terms of Mean-Per-Joint-Position-Error ( MPJPE ) by 7\% and 2\% on Human3.6M and MPI-INF-3DHP, respectively.

We further propose to learn a temporal motion model to exploit cues from sequential frames of a video to obtain anatomically coherent and smoothly varying poses, while preserving the realism across different activities. We show that a moving-window fully-connected network that takes previous $N$ poses performs extremely well at capturing temporal as well as anatomical cues from pose sequences. With the help of carefully designed controlled experiments we show the temporal and anatomical cues learned by the model to facilitate better understanding. We report an additional 7\% improvement on Human3.6M with the use of our temporal model and demonstrate real-time performance of the full pipeline at 30fps. Our final model improves the published state-of-the-art on Human3.6M~\cite{h36m_pami} and MPI-INF-3DHP~\cite{mono-3dhp2017} by 11.8\% and 12\%, respectively.

\vspace{-1em}
\section{Related Work} \label{sec:relatedWork}
This section presents a brief summary of the past work related to human pose estimation from three viewpoints: (1) ConvNet architectures and training strategies, (2) Utilizing structural constraints of human bodies, and (3) 3D pose estimation from video. The reader is referred to~\cite{SARAFIANOS20161} for a detailed review of the literature.

\textbf{ConvNet architectures:} Most existing ConvNet based approaches either directly regress 3D poses from the input image~\cite{Sun_2017_ICCV,li20143d,zhou2016deep,zhou2016sparseness} or infer 3D from 2D pose in a two-stage approach~\cite{Tome_2017_CVPR,Zhou_2017_ICCV,VNect_SIGGRAPH2017,Moreno-Noguer_2017_CVPR,Lin_2017_CVPR}. Some approaches make use of volumetric-heatmaps~\cite{Pavlakos_2017_CVPR}, some define a pose using bones instead of joints~\cite{Sun_2017_ICCV}, while the approach in~\cite{VNect_SIGGRAPH2017} directly regresses for 3D location maps. The use of 2D-to-3D pipeline enables training with large-scale in-the-wild 2D pose datasets~\cite{andriluka14cvpr,MSCOCO:2014}. A few approaches use statistical priors~\cite{zhou2016sparseness,akhter2015pose} to lift 2D poses to 3D. Chen et al.~\cite{Chen_2017_CVPR} and Yasin et al.~\cite{yasin2016dual} use a pose library to retrieve the nearest 3D pose given the corresponding 2D pose prediction. Recent ConvNet based approaches~\cite{VNect_SIGGRAPH2017,Rogez_2017_CVPR,Zhou_2017_ICCV,Sun_2017_ICCV,zhou2016sparseness,Pavlakos_2017_CVPR} have reported substantial improvements in real-world setting by pre-training or joint training of their 2D prediction modules, but it still remains an open problem.

\textbf{Utilizing structural information:} The structure of the human skeleton is constrained by fixed bone lengths, joint angle limits, and limb interpenetration constraints. Some approaches use these constraints to infer 3D from 2D joint locations. Akhter and Black~\cite{akhter2015pose} learn pose-dependent joint angle limits for lifting 2D poses to 3D via an optimization problem. Ramakrishna et al.~\cite{varunECCV2012} solve for anthropometric constraints in an activity-dependent manner. Recently, Moreno~\cite{Moreno-Noguer_2017_CVPR} proposed to estimate the 3D inter-joint distance matrix from 2D inter-joint distance matrix using a simple neural network architecture. These approaches do not make use of rich visual cues present in images and rely on the predicted 2D pose that leads to sub-optimal results. Sun et al.~\cite{Sun_2017_ICCV} re-parameterize the pose presentation to use bones instead of joints and propose a structure-aware loss. But, they do not explicitly seek to penalize the feasibility of inferred 3D pose in the absence of 3D ground-truth data. Zhou et al.~\cite{Zhou_2017_ICCV} introduce a weakly-supervised framework for joint training with 2D and 3D data with the help of a geometric loss function to exploit the consistency of bone-length ratios in human body. We further strengthen this weakly-supervised setup with the help of joint-angle limits and left-right symmetry based loss functions for better training. Lastly, there are methods that recover both shape and pose from a 2D image via a mesh-fitting strategy. Bogo et al.~\cite{bogo2016keep} penalize body-part interpenetration and illegal joint angles in their objective function for finding SMPL~\cite{DBLP:journals/tog/LoperM0PB15} based shape and pose parameters. These approaches are mostly offline in nature due to their computational requirements, while our approach runs at 30fps.

\textbf{Utilizing temporal information:} Direct estimation of 3D pose from disjointed images leads to temporally incoherent output with visible jitters and varying bone lengths. 3D pose estimates from a video can be improved by using simple filters or temporal priors. Mehta et al.~\cite{VNect_SIGGRAPH2017} propose a real-time approach which penalizes acceleration and depth velocity in an optimization step after generating 3D pose proposals using a ConvNet. They also smooth the output poses with the use of a tunable low-pass filter~\cite{casiez20121} optimized for interactive systems. Zhou et al.~\cite{zhou2016sparseness} introduce a first order smoothing prior in their temporal optimization step. Alldieck et al.~\cite{Alldieck2017} exploit 2D optical flow features to predict 3D poses from videos. Wei et al.~\cite{Wei:2010} exploit physics-based constraints to realistically interpolate 3D motion between video keyframes. There have also been attempts to learn motion models. Urtasun et al.~\cite{urtasun2006temporal} learn activity specific motion priors using linear models while Park et al.~\cite{Park:2006} use a motion library to find the nearest motion given a set of 2D pose predictions followed by iterative fine-tuning. The motion models are activity-specific whereas our approach is generic. Recently, Lin et al.~\cite{Lin_2017_CVPR} used recurrent neural networks to learn temporal dependencies from the intermediate features of their ConvNet based architecture. In a similar attempt, Coskun et al.~\cite{Coskun_2017_ICCV} use LSTMs to design a Kalman filter that learns human motion model. In contrast with the aforementioned approaches, our temporal model is simple yet effectively captures short-term interplay of past poses and predicts the pose of the current frame in a temporally and anatomically consistent manner. It is generic and does not need to be trained for activity-specific settings. We show that it learns complex, non-linear inter-joint dependencies over time; e.g. it learns to refine wrist position, for which the tracking is least accurate, based on the past motion of elbow and shoulder joints.

\section{Background and Notations} \label{sec:background}
This section introduces the notations used in this article and also provides the required details about the weakly-supervised framework of Zhou et al.~\cite{Zhou_2017_ICCV} for joint learning from 2D and 3D data. 

A 3D human pose \(P = \{p_1, p_2, \ldots, p_k \}\) is defined by the positions of $k$ = 16 body joints in Euclidean space. These joint positions are defined relative to a root joint, which is fixed as the pelvis. The input to the pose estimation system could be a single RGB image or a continuous stream of RGB images \(I = \{ \ldots, I_{i-1}, I_i\}\). The $i^{th}$ joint $p_i$ is the coordinate of the joint in a 3D Euclidean space i.e. $p_i = (p_i^x, p_i^y, p_i^z )$. Throughout this article inferred variables are denoted with a $\tilde{*}$ and ground-truth is denoted with a $\hat{*}$, therefore, an inferred joint will be denoted as $\tilde{p}$ and ground-truth as $\hat{p}$. The 2D pose can be expressed with only the x,y-coordinates and denoted as $p^{xy} = (p^x, p^y)$; the depth-only joint location is denoted as $p^z = (p^z)$. The $i^{th}$ training data from a 3D annotated dataset consists of an image $I_i$ and corresponding joint locations in 3D, $\hat{P}_i$. On the other hand, the 2D data has only the 2D joint locations, $\hat{P}_i^{xy}$. Armed with these notations, below we describe the weakly-supervised framework for joint learning from~\cite{Zhou_2017_ICCV}.

\begin{figure}[!h] 
	\centering
	\includegraphics[width=1\linewidth]{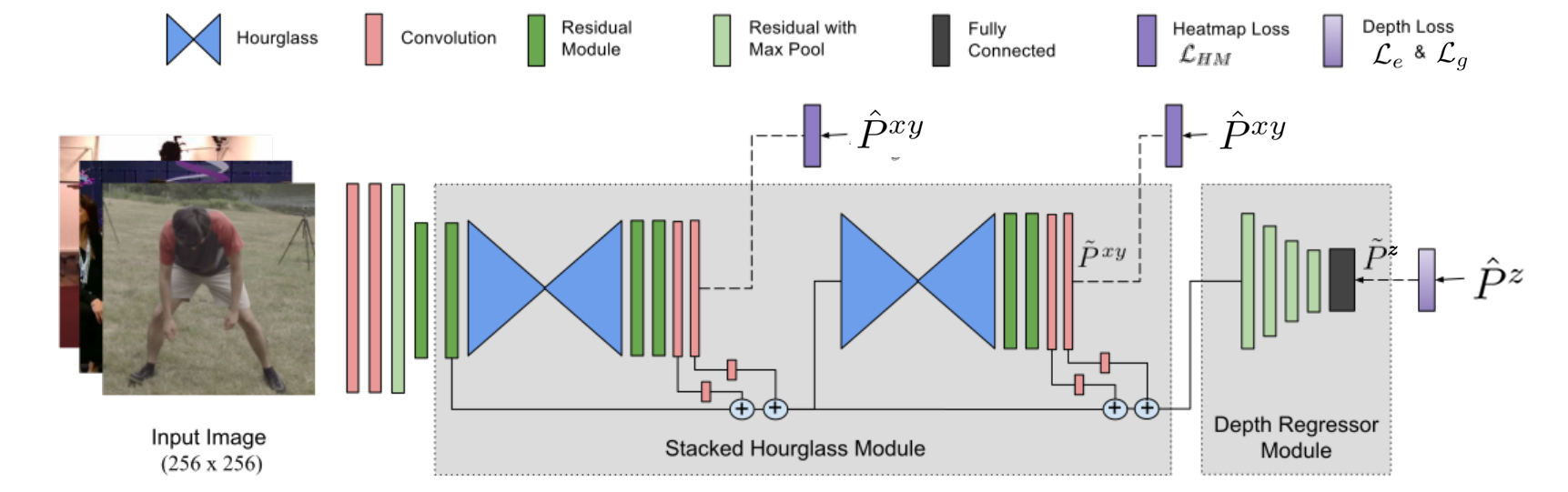}
    \caption{ A schematic of the network architecture. The stacked hourglass module is trained using the standard Euclidean loss $\mathcal{L}_{HM}$ against ground truth heatmaps. Whereas, the depth regressor module is trained on either $\mathcal{L}^z_{3D}$ or $\mathcal{L}^z_{2D}$ depending on whether the ground truth depth $\hat{P}^z$ is available or not.} 
	\vspace{-1em}
	\label{fig:architecture}
\end{figure}

Due to the absence of in-the-wild 3D data, the pose estimation systems learned using the controlled or synthetic 3D data fail to generalize well to in-the-wild settings. Therefore, Zhou et al.~\cite{Zhou_2017_ICCV} proposed a weakly-supervised framework for joint learning from both 2D and 3D annotated data. Joint learning exploits the 3D data for depth prediction and the in-the-wild 2D data for better generalization to real-world scenario. The overall schematic of this framework is shown in Fig.~\ref{fig:architecture}. It builds upon the stacked hourglass architecture~\cite{NewellYD16} for 2D pose estimation and adds a depth-regression sub-network on top of it. The stacked hourglass is trained to output the 2D joint locations, $\tilde{P}^{xy}$ in the image coordinate with the use of standard Euclidean loss between the predicted and the ground-truth joint-location heatmaps, please refer to~\cite{NewellYD16} for more details. The depth-regression sub-network, a series of four residual modules~\cite{he2016deep} followed by a fully connected layer, takes a combination of different feature maps from stacked hourglass and outputs the depth of each joint i.e. $\tilde{P}^z$. Standard Euclidean loss $\mathcal{L}_{e}(\tilde{P}^z, \hat{P}^z)$ is used for the 3D annotated data-sample. On the other hand, a weak-supervision in the form of a geometric loss function, $\mathcal{L}_{g}(\tilde{P}^{z}, \hat{P}^{xy})$, is used to train with a 2D-only annotated data-sample. The geometric loss acts as a regularizer and penalizes the pose configurations that violate the consistency of bone-length ratio priors. Please note that the ground-truth xy-coordinates, $\hat{P}^{xy}$, with inferred depth, $\tilde{P}^z$ are used in $\mathcal{L}_g$ to make the training simple.

The geometric loss acts as an effective regularizer for the joint training and improves the accuracy of 3D pose estimation under controlled and in-the-wild test conditions, but it ignores certain other \emph{strong} anatomical constraints of the human body. In the next section, we build upon the discussed weakly-supervised framework and propose a novel structure-aware loss that captures richer anatomical constraints and provides stronger weakly-supervised regularization than the geometric loss. 

\section{Proposed Approach}
This section introduces two novel anatomical loss functions and shows how to use them in the weakly-supervised setting to train with 2D annotated data-samples. Next, the motivation and derivation of the proposed losses and the analyses of the loss surfaces is presented to facilitate a deeper understanding and highlight the differences from the previous approaches. Lastly, a learnable temporal motion model is proposed with its detailed analysis through carefully designed controlled experiments.

\begin{figure}[!h] 
	\centering
	\includegraphics[width=1\linewidth]{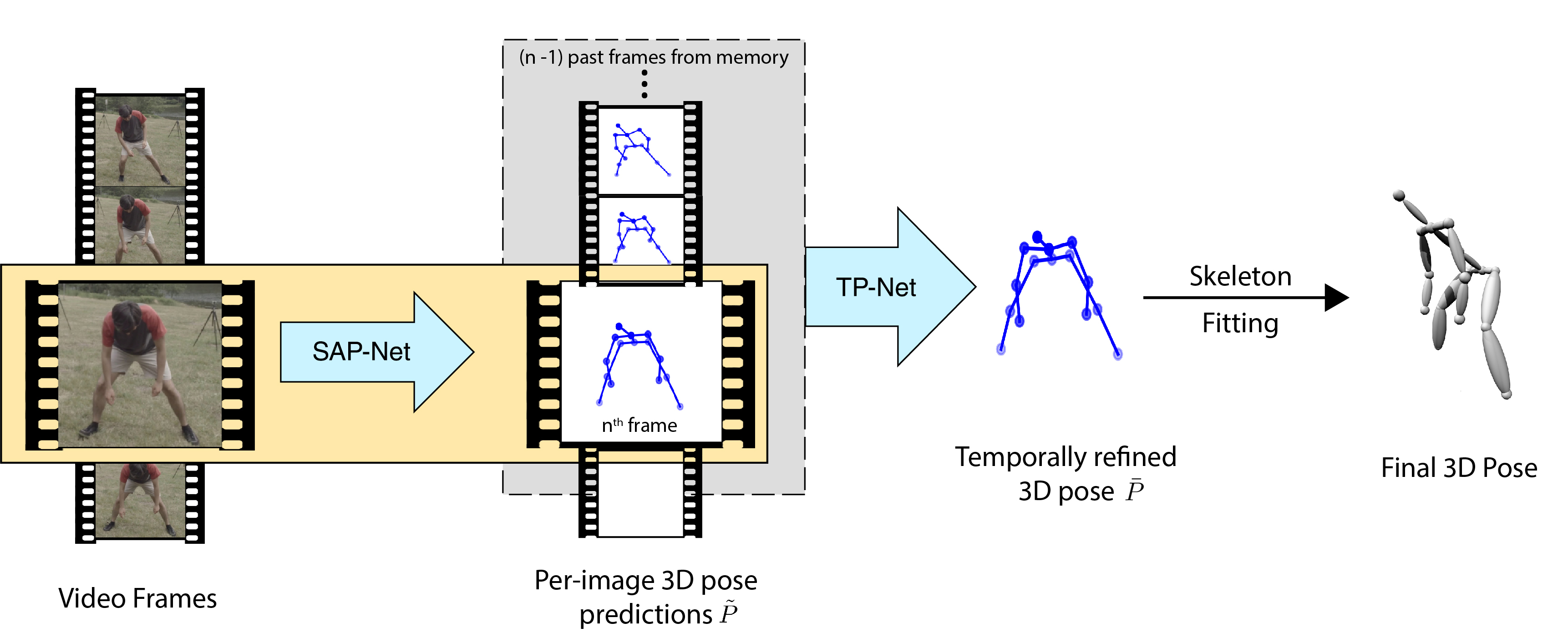}
    \caption{Overall pipeline of our method: We sequentially pass the video frames to a ConvNet that produces 3D pose outputs (one at a time). Next, the prediction is temporally refined by passing a context of past N frames along with the current frame to a temporal model. Finally, skeleton fitting may be performed as an optional step depending upon the application requirement.} 
	\vspace{-1em}
	\label{fig:pipeSchema}
\end{figure}

Fig.~\ref{fig:pipeSchema} shows our complete pipeline for 3D pose estimation. It consists of 
\begin{enumerate}
    \item \emph{\bf Structure-Aware PoseNet} or \emph{\bf SAP-Net}: A single-frame based 3D pose-estimation system that takes a single RGB image $I_i$ and outputs the inferred 3D pose $\tilde{P}_i$.
    
    \item \emph{\bf Temporal PoseNet} or \emph{\bf TP-Net}: A learned temporal motion model that can take a continuous sequence of inferred 3D poses $\{\ldots, \tilde{P}_{i-2}, \tilde{P}_{i-1}\}$ and outputs a temporally harmonized 3D pose $\Bar{P}_i$.
    
    \item \emph{\bf Skeleton fitting}: Optionally, if the actual skeleton information of the subject is also available, we can carry out a simple skeleton fitting step which preserves the directions of the bone vectors.
\end{enumerate}

\subsection{Structure-Aware PoseNet or SAP-Net}
SAP-Net uses the network architecture shown in Fig.~\ref{fig:pipeSchema}, which is taken from ~\cite{Zhou_2017_ICCV}. This network choice allows joint learning with both 2D and 3D data in weakly-supervised fashion as described in Section~\ref{sec:background}. A 3D annotated data-sample provides strong supervision signal and drives the inferred depth towards a unique solution. On the other hand, weak-supervision, in the form of anatomical constraints, imposes penalty on invalid solutions, therefore, restricts the set of solutions. Hence, the stronger and more comprehensive the set of constraints, the smaller and better the set of solutions. We seek motivation from the discussion above and propose to use loss functions derived from joint-angle limits and left-right symmetry of human body in addition to bone-length ratio priors~\cite{Zhou_2017_ICCV} for weak-supervision. Together, these three constraints are stronger than the bone-length ratio prior only and lead to better 3D pose configurations. For example, bone-length ratio prior will consider an elbow bent backwards as valid, if the bone ratios are not violated, but the joint-angle limits will invalidate it. Similarly, the symmetry loss eliminates the configurations with asymmetric left-right halves in the inferred pose. Next we describe and derive differentiable loss functions for the proposed constraints. 

\vspace{-1em}
\subsubsection{Illegal Angle Loss ($\mathcal{L}_a$):} Most body joints are constrained to move within a certain angular limits only. Our illegal angle loss,  $\mathcal{L}_a$, encapsulates this constraint for the knee and elbow joints and restricts their bending beyond $180^{\circ}$. For a given 2D pose $P^{xy}$, there exist multiple possible 3D poses and $\mathcal{L}_a$ penalizes the 3D poses that violate the knee or elbow joint-angle limits. To exploit such constraints, some methods ~\cite{HERDA2005189,akhter2015pose,ChenNie2013TIP} use non-differentiable functions to infer the legality of a pose. Unfortunately, the non-differentiability restricts their direct use in training a neural network. 
Other methods resort to represent a pose in terms of rotation matrices or quarternions for imposing joint-angle limits ~\cite{akhter2015pose,Wei:2010} that affords differentiability. However, this imposition is non-trivial when representing poses in terms of joint-positions, which are a more natural representation for ConvNets. 

\begin{figure}[!tb] 
	\centering
	\includegraphics[width=0.4\linewidth]{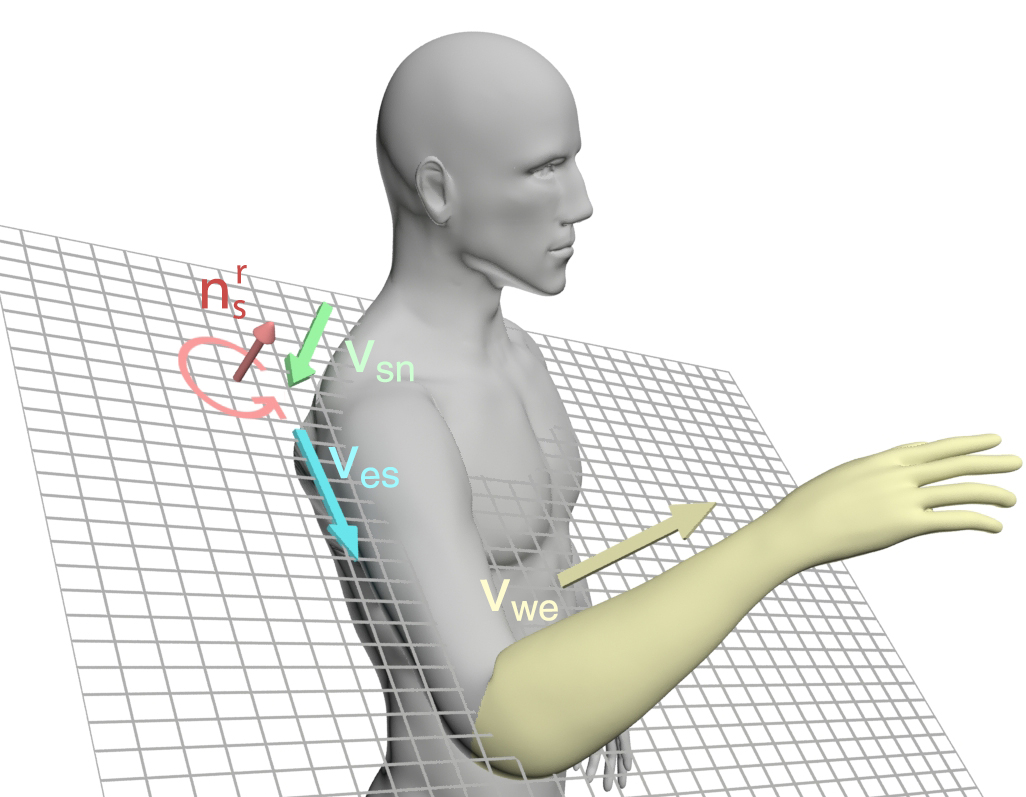}
    \caption{Illustration of Illegal Angle loss: For the elbow joint angle to be legal, the lower-arm must project a positive component along $\mathbf{n_s^r}$ (normal to collarbone-upperarm plane) , i.e. $\mathbf{n_s^r} \cdot \mathbf{v_{we}} \geq 0$. Note that we only need 2D annotated data to train our model using this formulation.}
    \vspace{-1em}
    \label{fig:angLoss}
\end{figure}

Our novel formulation of illegal-angle discovery resolves the ambiguity involved in differentiating between the internal and external angle of a joint for a 3D joint-location based pose representation. Using our formulation and keeping in mind our the requirement of differentiability, we formulate $\mathcal{L}_a$ to be used directly as a loss function.  We illustrate our formulation with the help of Fig.~\ref{fig:angLoss}, and explain its derivation for the right elbow joint. Subscripts $n$, $s$, $e$, $w$, $k$ denote neck, shoulder, elbow, wrist and knee joints in that order, and superscripts $l$ and $r$ represent left and right body side, respectively. 
We define \(\mathbf{v_{sn}^r} = P_s^r - P_n\), \(\mathbf{v_{es}^r} = P_e^r - P_s^r \) and \(\mathbf{v_{we}^r} = P_w^r - P_e^r \) as the collar-bone, upper-arm and the lower-arm, respectively (See Fig.~\ref{fig:angLoss}). Now, \(\mathbf{n_s^r} = \mathbf{v_{sn}^r \times \mathbf{v_{es}^r}  }\) is the normal to the plane defined by the collar-bone and the upper-arm. For the elbow joint to be legal, \(\mathbf{v_{we}^r}\) must have a positive component in the direction of $\mathbf{n_s^r}$, i.e. \(\mathbf{n_s^r} \cdot \mathbf{v_{we}^r} \) must be positive. We do not incur any penalty when the joint angle is legal and define \(E_e^r = \min(\mathbf{n_s^r} \cdot \mathbf{v_{we}^r}, 0)\) as a measure of implausibility. Note that this case is opposite for the right knee and left elbow joints (as shown by the right hand rule) and requires $E_k^r$ and $E_e^l$ to be positive for the illegal case. We exponentiate $E$ to strongly penalize large deviations beyond legality. $\mathcal{L}_a$ can now be defined as: \\ 

\begin{equation} \label{eq:langle}
 \mathcal{L}_a = -E_e^r e^{-E_e^r} + E_e^l e^{E_e^l} + E_k^r e^{E_k^r} - E_k^l e^{-E_k^l} 
\end{equation}

All the terms in the loss are functions of bone vectors which are, in turn, defined in terms of the inferred pose. Therefore, $\mathcal{L}_a$ is differentiable. Please refer to the supplementary material for more details. 

\vspace{-1em}
\subsubsection{Symmetry Loss ($\mathcal{L}_s$):} It is simple yet heavily constrains the joint depths, especially when the inferred depth is ambiguous due to occlusions. $\mathcal{L}_s$ is defined as the difference in lengths of left/right bone pairs. Let $\mathcal{B}$ be the set of all the bones on right half of the body except torso and head bones. Also, let $BL_b$ represent the bone-length of bone $b$. We define $L_s$ as\\
\begin{equation}
\mathcal{L}_s = \sum_{b \in \mathcal{B}}  \vert\vert{ BL_b - BL_{C(b)}}\vert\vert_2 
\end{equation} 
where $C(.)$ indicates the corresponding left side bone.

Finally, our structure-aware loss $\mathcal{L}^z_{SA}$ is defined as weighted sum of illegal-angle loss $\mathcal{L}^z_{a}$, symmetry-loss $\mathcal{L}^z_{s}$ and geometric loss $\mathcal{L}^z_{g}$ from~\cite{Zhou_2017_ICCV} - 
\begin{equation} \label{eq:l2d}
\begin{split}
\mathcal{L}^z_{SA}(\tilde{P}^z, \hat{P}^{xy}) & =  \lambda_{a} \mathcal{L}_{a}(\tilde{P}^z, \hat{P}^{xy}) + \lambda_{s} \mathcal{L}_{s}(\tilde{P}^z, \hat{P}^{xy})  + \lambda_g\mathcal{L}_{g}(\tilde{P}^z, \hat{P}^{xy})
  \end{split}
\end{equation}
\vspace{-3em}

\subsubsection{Loss Surface Visualization:} 
\begin{figure*}[!h] 
	\centering
	\includegraphics[width=1\textwidth]{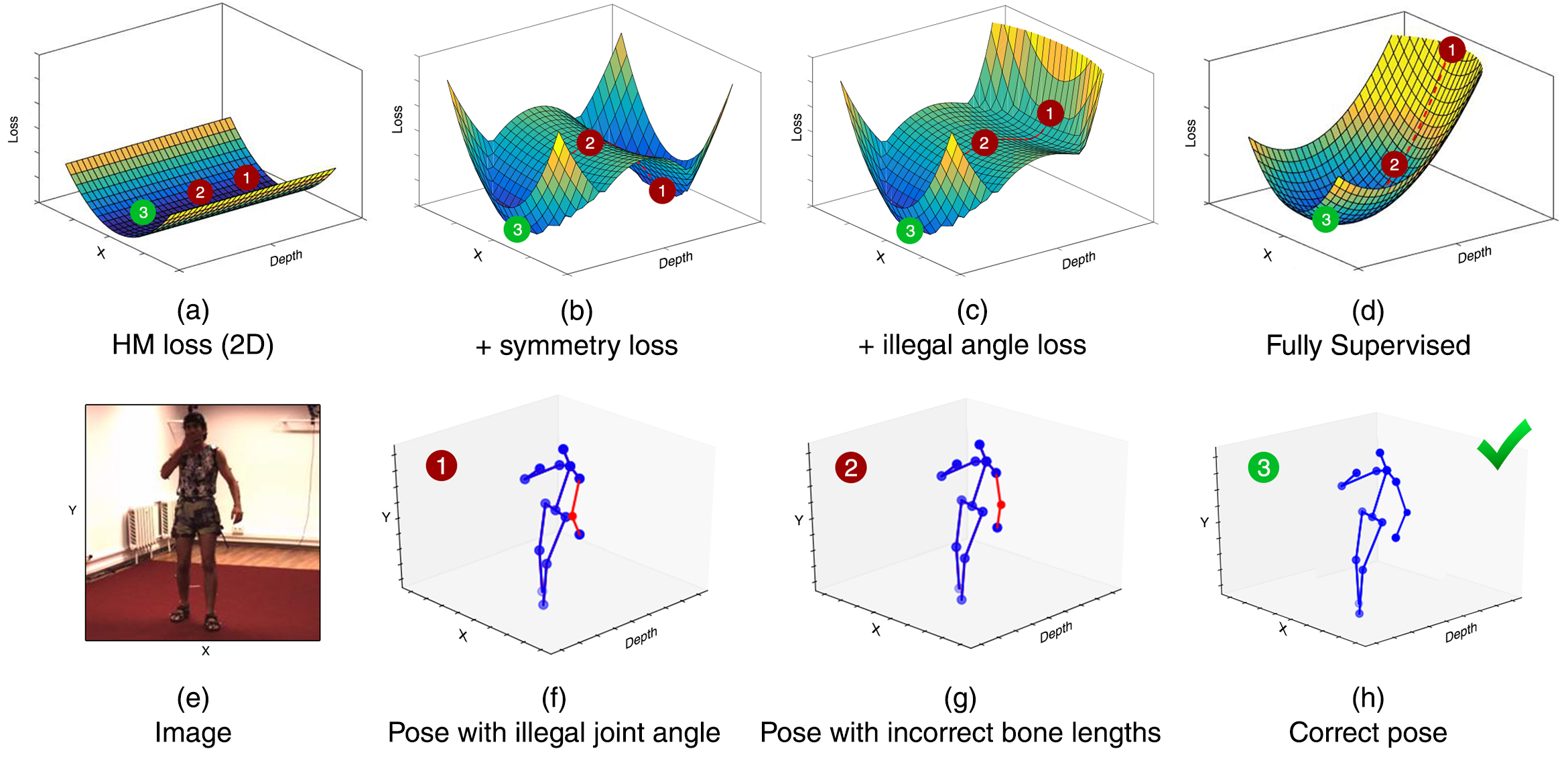}
    \caption{\textbf{Loss Surface Evolution} Plots (a) to (d) show the local loss surfaces for (a) 2D-location loss. (b) 2D-location+symmetry loss (c) 2D-location+symmetry+illegal angle loss and (d) full 3D-annotation Euclidean loss. The points (1), (2) and (3) highlighted on the plots are the corresponding 3D poses shown in (f), (g) and (h), with (3) being the ground-truth depth. The illegal angle penalty increases the loss for pose (1), which has the elbow bent backwards. Pose (2) has a legal joint angle, but the symmetry is lost. Pose (3) is correct. We can see that without the angle loss, the loss at (1) and (3) are equal and we cannot discern between the two points.} 
    \vspace{-1em}
    \label{fig:surfEnergy}
\end{figure*}

Here we take help of local loss surface visualization to appreciate how the proposed losses are pushing invalid configurations towards their valid counterparts. In order to obtain the loss surfaces we take a random pose $P$ and vary the $(x_{le},z_{le})$ coordinates of left elbow over an $XZ$ grid while keeping all other joint locations fixed. Then, we evaluate $\mathcal{L}^z_{SA}$ at different $(x,z)$ locations in the $XZ$ grid to obtain the loss, which is plotted as surfaces in Fig.~\ref{fig:surfEnergy}. We plot loss surfaces with only 2D-location loss, 2D-location+symmetry loss, 2D-location+symmetry+illegal angle loss and 3D-annotation based Euclidean loss to show the evolution of the loss surfaces under different anatomical constraints. From the figure it is clear that both the symmetry loss and illegal angle loss morph the loss surface to facilitate moving away from illegal joint configurations.

\vspace{-2em}
\subsection{Temporal PoseNet or TP-Net}
\begin{figure*}[!h]
	\centering
	\includegraphics[width = \textwidth]{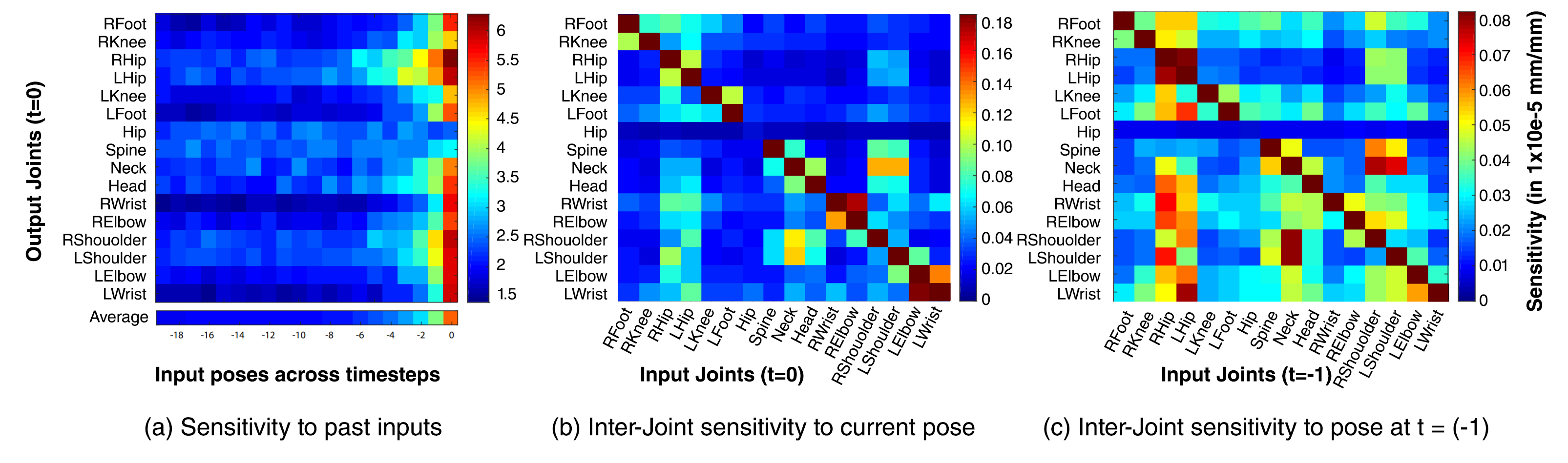}
    \caption{(a) The variation of sensitivity in output pose w.r.t to the perturbations in input poses of TP-Net for from $t$=0 to $t$=-19. (b) Strong structural correlations are learned from the pose input at $t$=0 frame. (c) Past frames show smaller but more complex structural correlations. The self correlations (diagonal elements) are an order of magnitude larger and the colormap range has been capped to better display.
    }
    \label{fig:tempInf}
    \vspace{-2em}
\end{figure*}

In this section we propose to learn a temporal pose model, referred as Temporal PoseNet, to exploit the temporal consistency and motion cues present in video sequences. Given independent pose estimates from SAP-Net, we seek to exploit the information from a set of adjacent pose-estimates $\mathbf{P_{adj}}$ to improve the inference for the required pose $P$. We propose to use a simple two-layer, 4096 hidden neurons, fully-connected network with ReLU non-linearity that takes a fixed number, $N=20$, of adjacent poses as inputs and outputs the required pose $\Bar{P}$. The adjacent pose vectors are simply flattened and concatenated in order to make a single vector that goes into the TP-Net and it is trained using standard $L_2$ loss from the ground-truth pose. Despite being extremely simple in nature, we show that it outperforms a more complex variant such as RNNs, see Table~\ref{tab:lstmComp}. Why? We believe it happens because intricate human motion has increasing variations possible with increasing time window, which perhaps makes additional information from too far in the time useless or at least difficult to utilize. Therefore, a dense network with a limited context can effectively capture the useful consistency and motion cues. 

In order to visualize the temporal and structural information exploited by TP-Net we carried out a simple sensitivity analysis in which we randomly perturbed the joint locations of $P_t$ that is $t$ time-steps away from the output of TP-Net $\Bar{P}$ and plot the sensitivity for time-steps $t=-1$ to $t=-19$ for all joints in Fig.~\ref{fig:tempInf}(a). We can observe that poses beyond 5 time-steps ( or $200ms$ time-window ) does not have much impact on the predicted pose. Similarly, Fig.~\ref{fig:tempInf}(b) shows the structural correlations the model has learned just within the current frame. TP-Net learns to rely on the locations of hips and shoulders to refine almost all the other joints. We can also observe that the child joints are correlated with parent joints, for eg. the wrists are strongly correlated with elbows, and the shoulders are strongly correlated with the neck. Fig.~\ref{fig:tempInf}(c) shows the sensitivity to the input pose at $t$ = -1. Here, the correlations learned from the past are weak, but exhibit a richer pattern. The sensitivity of the child joints extends further upwards into the kinematic chain, eg. the wrist shows higher correlations with elbow, shoulder and neck, for the $t$ = -1 frame. Therefore, we can safely conclude that TP-Net learns complex structural and motion cues despite being so simple in nature. We hope this finding would be useful for future research in this direction

Since TP-Net takes as input a fixed number of adjacent poses, we can choose to take all the adjacent poses before the required pose, referred to as \emph{online} setting, or we can choose to have $N/2=10$ adjacent poses on either side of required pose, referred to as \emph{semi-online} setting. Since our entire pipeline runs at 30fps, even semi-online setting will run at a lag of 10fps only. From Fig.~\ref{fig:tempInf} we observe that TP-Net can learn complex, non-linear inter-joint dependencies over time - for e.g. it learns to refine wrist position, for which the tracking is least accurate, based on the past motion of elbow and shoulder joints.

\subsection{Training and Implementation details} \label{sec:training}
While training the SAP-Net, both 2D samples, from MPII2D, and 3D samples, from either of the 3D datasets, were consumed in equal proportion in each iteration with a minibatch size of 6. In the \emph{first stage} we obtain a strong 2D pose estimation network by pre-training the hourglass modules of SAP-Net on MPII and Human3.6 using SGD as in~\cite{NewellYD16}. Training with weakly-supervised losses require a warm start~\cite{zhou2017brief}, therefore, in the \emph{second stage} we train the 3D depth module with only 3D annotated data-samples for 240k iterations so that it learns to output reasonable poses before switching on weak-supervision. In the \emph{third stage} we train SAP-Net with $\mathcal{L}_g$ and $\mathcal{L}_a$ for 160k iterations with $\lambda_a = 0.03$, $\lambda_g = 0.03$ with a learning-rate of $2.5e-4$. Finally, in the \emph{fourth stage} we introduce the symmetry loss, $\mathcal{L_s}$ with $\lambda_s = 0.05$ and learning-rate $2.5e-5$. 

TP-Net was trained using Adam optimizer~\cite{kingma2014adam} for 30 epochs using the pose predictions generated by fully-trained SAP-Net. 
In our experiments, we found that a context of $N = 20$ frames yields the best improvement on MPJPE (Fig.~\ref{fig:tempInf}) and we use that in all our experiments. It took approximately two days to train SAP-Net and one hour to train TP-Net using one NVIDIA 1080 Ti GPU. SAP-Net runs at an average testing time of $20ms$ per image while TP-Net adds negligible delay (\textless1ms).

\begin{table*}[t]
\centering
\begin{tabular}{l  c  c  c  c  c  c  c  c }
\hline
Method & Direction & Discuss & Eat & Greet & Phone & Pose & Purchase & Sit \\
\hline
\hline
Zhou~\cite{zhou2016sparseness} & 68.7 & 74.8 & 67.8 & 76.4 & 76.3 & 84.0 & 70.2 & 88.0 \\
Jahangiri~\cite{Jahangiri:ICCV2017} & 74.4 & 66.7 & 67.9 & 75.2 & 77.3 & 70.6 & 64.5 & 95.6 \\
Lin~\cite{Lin_2017_CVPR} & 58.0 & 68.2 & 63.2 & 65.8 & 75.3 & 61.2 & 65.7 & 98.6 \\
Mehta~\cite{mono-3dhp2017}  & 57.5 & 68.6 & 59.6 & 67.3 & 78.1 & 56.9 & 69.1 & 98.0 \\
Pavlakos~\cite{Pavlakos_2017_CVPR} & 58.6 & 64.6 & 63.7 & 62.4 & 66.9 & 57.7 & 62.5 & 76.8 \\
Zhou~\cite{Zhou_2017_ICCV} & 54.8 & 60.7 & 58.2 & 71.4 & 62.0 & 53.8 & 55.6 & 75.2 \\
Sun~\cite{Sun_2017_ICCV} & 52.8 & 54.8 & 54.2 & 54.3 & 61.8 & 53.1 & 53.6 & 71.7 \\
\hline
Ours(SAP-Net) & 46.9 & 53.8 & 47.0 & 52.8 & 56.9 & 45.2 & 48.2 & 68.0  \\
Ours(TP-Net) & \textbf{44.8} & \textbf{50.4} & \textbf{44.7} & \textbf{49.0} & \textbf{52.9} & \textbf{43.5} & \textbf{45.5} & \textbf{63.1} \\

\hline
\hline
Method & SitDown & Smoke & Photo & Wait & Walk & WalkDog & WalkPair & Avg \\
\hline
\hline
Zhou~\cite{zhou2016sparseness} & 113.8 & 78.0 & 78.4 & 89.1 & 62.6 & 75.1 & 73.6 & 79.9 \\
Jahangiri~\cite{Jahangiri:ICCV2017} & 127.3 & 79.6 & 79.1 & 73.4 & 67.4 & 71.8 & 72.8 & 77.6 \\
Lin~\cite{Lin_2017_CVPR} & 127.7 & 70.4 & 93.0 & 68.2 & 50.6 & 72.9 & 57.7 & 73.1 \\
Mehta~\cite{mono-3dhp2017} & 117.5 & 69.5 & 82.4 & 68.0 & 55.3 & 76.5 & 61.4 & 72.9 \\
Pavlakos~\cite{Pavlakos_2017_CVPR} & 103.5 & 65.7 & 70.7 & 61.6 & 56.4 & 69.0 & 59.5 & 66.9 \\
Zhou~\cite{Zhou_2017_ICCV}& 111.6 & 64.1 & 65.5 & 66.0 & 51.4 & 63.2 & 55.3 & 64.9 \\
Sun~\cite{Sun_2017_ICCV}& \textbf{86.7} & 61.5 & 67.2 & 53.4 & 47.1 & 61.6 & 53.4 & 59.1 \\
\hline
Ours(SAP-Net) & 94.0 & 55.7 & 63.6 & 51.6 & 40.3 & 55.4 & 44.3 & 55.5 \\
Ours(TP-Net) & 87.3 & \textbf{51.7} & \textbf{61.4} & \textbf{48.5} & \textbf{37.6} & \textbf{52.2} & \textbf{41.9} & \textbf{52.1} \\
\hline
\end{tabular}
\vskip 2mm
\caption{Comparative evaluation of our model on Human 3.6 following Protocol 1. The evaluations were performed on subjects 9 and 11 using ground truth bounding box crops and the models were trained only on Human3.6 and MPII 2D pose datsets.} \label{tab: h36mp1}
\vspace{-2em}
\end{table*}

\begin{table*}[!h]
\fontsize{7}{8}\selectfont 
\centering
\setlength\tabcolsep{1pt}
\begin{tabular}{lcccccccccccccccc}
\hline
Method & Direct. & Discuss & Eat & Greet & Phone & Pose & Purch. & Sit & \shortstack{Sit\\Down} & Smoke & Photo & Wait & Walk & \shortstack{Walk\\Dog} & \shortstack{Walk\\Pair} & Avg  \\
\hline
\hline
Yasin~\cite{yasin2016dual} & 88.4 & 72.5 & 108.5 & 110.2 & 97.1 & 91.6 & 107.2 & 119.0  & 170.8 & 108.2 & 142.5 & 86.9 & 92.1 & 165.7 & 102.0 & 108.3 \\
Rogez~\cite{rogezNIPS} & - & - & - & - & - & - & - & - & - & - & - & - & - & - & - & 88.1 \\
Chen~\cite{Chen_2017_CVPR} & 71.6 & 66.6 & 74.7 & 79.1 & 70.1 & 67.6 & 89.3 & 90.7  & 195.6 & 83.5 & 93.3 & 71.2 & 55.7 & 85.9 & 62.5 & 82.7  \\
Nie~\cite{Nie_2017_ICCV} & 62.8 & 69.2 & 79.6 & 78.8 & 80.8 & 72.5 & 73.9 & 96.1 & 106.9 & 88.0 & 86.9 & 70.7 & 71.9 & 76.5 & 73.2 & 79.5  \\
Moreno~\cite{Moreno-Noguer_2017_CVPR} & 67.4 & 63.8 & 87.2 & 73.9 & 71.5 & 69.9 & 65.1 & 71.7  & 98.6 & 81.3 & 93.3 & 74.6 & 76.5 & 77.7 & 74.6 & 76.5 \\
Zhou~\cite{zhou2016sparseness} & 47.9 & 48.8 & 52.7 & 55.0 & 56.8 & 49.0 & 45.5 & 60.8 & 81.1 & 53.7 & 65.5 & 51.6 & 50.4 & 54.8 & 55.9 & 55.3  \\
Sun~\cite{Sun_2017_ICCV} & 42.1 & 44.3 & 45.0 & 45.4 & 51.5 & 43.2 & 41.3 & 59.3 & 73.3 & 51.0 & 53.0 & 44.0 & 38.3 & 48.0 & 44.8 & 48.3 \\
\hline
Ours(SAP-Net) & 32.8 & 36.8 & 42.5 & 38.5 & 42.4 & 35.4 & 34.3 &  53.6 & 66.2 & 46.5 & 49.0 & 34.1 & 30.0 & 42.3 & 39.7 & 42.2  \\
Ours (TP-Net) & \textbf{28.0} & \textbf{30.7} & \textbf{39.1} & \textbf{34.4} & \textbf{37.1} & \textbf{28.9} & \textbf{31.2} & \textbf{39.3} & \textbf{60.6} & \textbf{39.3} & \textbf{44.8} & \textbf{31.1} & \textbf{25.3} & \textbf{37.8} & \textbf{28.4} & \textbf{36.3}  \\
\hline
\end{tabular}
\vskip 2mm
  \caption{Comparative evaluation of our model on Human 3.6M using Protocol 2. The models were trained only on Human3.6M and MPII 2D datasets.} \label{tab:h36mp2}
\vspace{-2em}
\end{table*}

\begin{table}[!bht]
\centering
\footnotesize
\parbox{.40\linewidth}{

\begin{tabular}{ l  l  c  c c  }
{\bf Method} & {\bf MPJE}\\
\hline
Zhou w/o $\mathcal{L}_g$~\cite{Zhou_2017_ICCV}& 65.69\\
 + Geometry loss & 64.90\\
\hline
Baseline & 58.50\\
 + Geometry loss & 58.45\\ 
 + Illegal Angle loss & 56.20\\
 + Symmetry loss & 55.51\\
 + TP-Net real-time & 52.10\\
 + TP-Net bi-directional & \textbf{51.10}\\
\hline
\end{tabular} 
\vskip 2mm
\caption{Ablation of different loss terms on Human3.6M using Protocol 1.}
\label{tab:ablation}
}
\hspace{1em}
\parbox{.45\linewidth}{

  \begin{tabular}{l  l  c  c c  }
     {\bf Model} &
      \multicolumn{3}{c}{\bf Number of input frames} \\
      \hline
      & 4 & 10 & 20\\
    \hline
    LSTM & - & - & 54.05 \\
    Bi-LSTM & 53.86 & 53.72 & 53.65 \\
    TP-Net (Ours) & 53.0 & 52.24 & 52.1 \\
    Bi-TP-Net (Ours) & 52.4 & 51.36 & \textbf{51.1} \\
    \hline
  \end{tabular}
\vskip 2mm
\caption{Comparison of different temporal models considered with varying context sizes. LSTM nets model the entire past context till time t. Bidirectional networks take half contextual frames from the future and half from the past.}
\label{tab:lstmComp}
}\\
\end{table}


\section{Experiments} \label{experiments}
In this section, we present ablation studies, quantitative results on Human3.6M and MPI-INF-3DHP datasets and comparisons with previous art, and qualitative results on MPII 2D and MS COCO datasets. We start by describing the datasets used in our experiments.

\indent \textbf{Human3.6M} has $11$ subjects performing different indoor actions with ground-truth annotations captured using a marker-based MoCap system. We follow ~\cite{Tome_2017_CVPR} and evaluate our results under 1) \textit{Protocol 1} that uses Mean Per Joint Position Error (MPJPE) as the evaluation metric w.r.t. root relative poses and 2) \textit{Protocol 2} that uses Procrustes Aligned MPJPE (PAMPJPE) which is MPJPE calculated after rigid alignment of predicted pose with the ground truth.

\textbf{MPI-INF-3DHP (test) dataset} is a recently released dataset of $6$ test subjects with different indoor settings ( green screen and normal background) and $2$ subjects performing in-the-wild that makes it more challenging than Human3.6M, which only has a single indoor setting. We follow the evaluation metric proposed in~\cite{mono-3dhp2017} and report Percentage of Correct Keypoints (PCK) within \textit{150mm} range and Area Under Curve (AUC). Like~\cite{Zhou_2017_ICCV}, we assume that the global scale is known and perform skeleton retargeting while training to account for the difference of joint definitions between Human3.6M and MPI-INF-3DHP datasets. Finally, skeleton fitting is done as an optional step to fit the pose into a skeleton of known bone lengths.

\begin{figure*}[t]
	\centering
	\includegraphics[width = \linewidth]{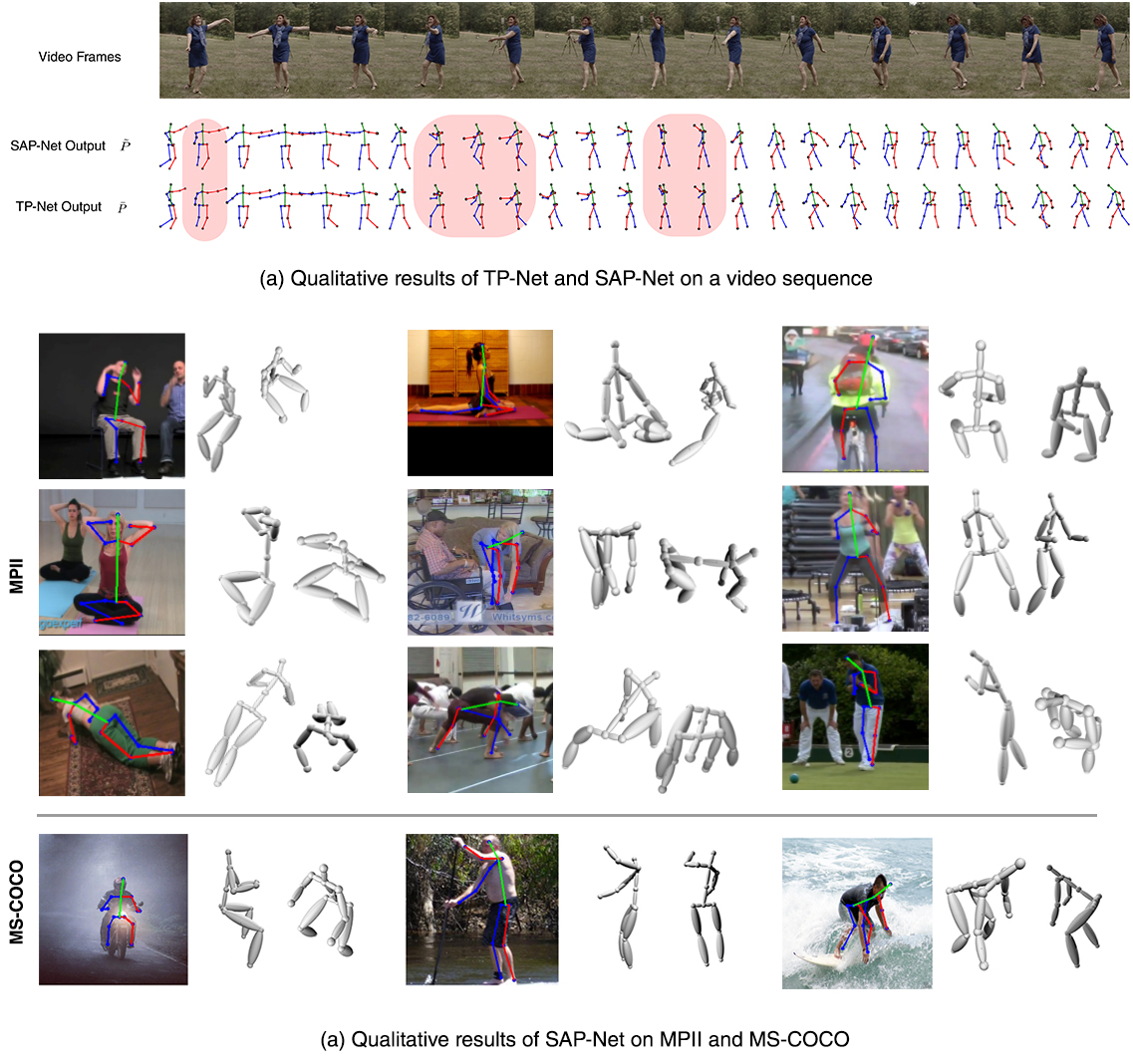}
    \caption{(a) Comparison of our temporal model TP-Net with SAP-Net on a video. The highlighted poses demonstrate the ability of TP-Net to learn temporal correlations, and smoothen and refine pose estimates from SAP-Net. (b) Qualitative results of SAP-Net on some images from MPII and MS-COCO datasets, from multiple viewpoints.}
    \label{fig:coco_vis}
    \vspace{-2em}
\end{figure*}

\textbf{2D datasets:} MS-COCO and MPII are in-the-wild 2D pose datasets with no 3D ground truth annotations. Therefore, we show qualitative results for both of them in Fig. ~\ref{fig:coco_vis}. Despite lack of depth annotation, our approach generalizes well and predicts valid 3D poses under background clutter and significant occlusion. 

\vspace{-1em}

\subsection{Quantitative Evaluations} \label{quanteval}
We evaluate the outputs of the three stages of our pipeline and show improvements at each stage.
\begin{enumerate}
    \item {\bf Baseline}: We train the same network architecture as SAP-Net but with only the fully supervised losses i.e. 2D heatmap supervision and $\mathcal{L}^e$ for 3D data only. 
    \item  {\bf SAP-Net}: Trained with the proposed structure-aware loss following Section~\ref{sec:training}.
    \item {\bf TP-Net}: Trained on the outputs of SAP-Net from video sequences ( see Section~\ref{sec:training}).
    \item {\bf Skeleton Fitting (optional)}: We fit a skeleton based on the subject's bone lengths while preserving the bone vector directions obtained from the 3D pose estimates.
\end{enumerate}
Below, we conduct ablation study on SAP-Net and report results on the two datasets.

\textbf{SAP-Net Ablation Study:} In order to understand the effect of individual anatomical losses, we train SAP-Net with successive addition of geometry $\mathcal{L}^z_{g}$, illegal-angle $\mathcal{L}^z_{a}$ and symmetry $\mathcal{L}^z_{s}$ losses and report their performance on Human3.6M under {\it Protocol 1} in Table~\ref{tab:ablation}. We can observe that the incorporation of illegal-angle and symmetry losses to geometry loss significantly improves the performance while geometry loss does not offer much improvement even over the baseline. Similarly, TP-Net offers significant improvements over SAP-Net and the \emph{semi-online} variant of TP-Net ( TP-Net bi-directional ) does even better than TP-Net.

\textbf{Evaluations on Human3.6M:}
We show significant improvement over the state-of-the-art and achieve an MPJPE of $55.5mm$ with SAP-Net which is further improved by TP-Net to $52.1mm$. Table~\ref{tab: h36mp1} and Table~\ref{tab:h36mp2} present a comparative analysis of our results under \textit{Protocol 1} and \textit{Protocol 2}, respectively. We outperform other competitive approaches by significant margins leading to an improvement of 12\%. 

\textbf{Evaluations on MPI-INF-3DHP:}
The results from Table~\ref{tab:mpi_full} show that we achieve slightly worse performance in terms of PCK and AUC but much better performance in terms of MPJPE, improvement of 12\%, as compared to the current state-of-the-art. It is despite the lack of data augmentation through green-screen compositing during training. 

\begin{table*}[th]
\centering
\parbox{.36\linewidth}{
\begin{tabular}{ l  c  c  c}
\hline
{\bf Method} & \shortstack{{\bf PCK}} & \shortstack{{\bf AUC}} & \shortstack{ {\bf MPJPE} } \\
\hline
Mehta~\cite{mono-3dhp2017} & 75.7 & 39.3 & 117.6 \\
Mehta~\cite{VNect_SIGGRAPH2017} & 76.6 & \textbf{40.4} & 124.7 \\
\hline
Ours & \textbf{76.7} & 39.1 & \textbf{103.8} \\
\hline
\end{tabular}
\vskip 2mm
\small
\caption{Results on MPI-INF-3DHP dataset. Higher PCK and AUC are desired while a lower MPJPE is better. Note that unlike ~\cite{mono-3dhp2017,VNect_SIGGRAPH2017}, the MPI-INF-3DHP training dataset was not augmented.}\label{tab:mpi_full}
\vspace{-1em}
}
\hspace{1em}
\parbox{.55\linewidth}{
\centering
\footnotesize
\begin{tabular}{ l c  c  c  c }
{\bf Bone} & {\bf Zhou~\cite{Zhou_2017_ICCV}} & {\bf SAP-Net} & {\bf TP-Net} \\
\hline
Upper arm & 37.8 & $25.8_{\downarrow31.7\%}$ & $\textbf{23.9}_{\downarrow36.7\%}$  \\
Lower arm & 50.7 & $\textbf{32.1}_{\downarrow36.7\%}$ & $33.9_{\downarrow33.1\%}$ \\
Upper leg & 43.4 & $27.8_{\downarrow35.9\%}$ & $\textbf{24.8}_{\downarrow42.8\%}$ \\
Lower leg & 47.8 & $38.2_{\downarrow20.1\%}$ & $\textbf{29.2}_{\downarrow38.9\%}$ \\
\hline
\hline
Upper arm & -- & 49.6 & \textbf{39.8} \\
Lower arm & -- & 66.0 & \textbf{48.3} \\
Upper leg & -- & 61.3 & \textbf{48.8} \\
Lower leg & -- & 68.8 & \textbf{48.3}  \\
\hline
\end{tabular} 
\vskip 2mm
\caption{Evaluating our models on (i) symmetry - mean $L_1$ distance in mm between left/right bone pairs (upper half), and (ii) the standard deviation (in mm) of bone lengths across all video frames (lower half) on MPI-INF-3DHP dataset.} \label{tab:symmetryTab}
\vspace{-1em}
}
\end{table*}

\vspace{-1em}
\subsection{Structural Validity Analysis}
This section analyzes the validity of the predicted 3D poses in terms of the anatomical constraints, namely left-right symmetry and joint-angle limits. Ideally, the corresponding left-right bone pairs should be of similar length; therefore, we compute the mean $L_1$ distance in mm between the corresponding left-right bone pairs on MPI-INF-3DHP dataset and present the results in the upper half of Table~\ref{tab:symmetryTab}. For fairness of comparison, we evaluate on model trained only on Human3.6M. We can see that SAP-Net, trained with symmetry loss, significantly improves the symmetry as compared to the system in~\cite{Zhou_2017_ICCV} which uses bone-length ratio priors and TP-Net offers further improvements by exploiting the temporal cues from adjacent frames. It shows the importance of explicit enforcement of symmetry. Moreover, it clearly demonstrates the effectiveness of TP-Net in implicitly learning the symmetry constraint. The joint-angle validity of the predicted poses is evaluated using~\cite{akhter2015pose} and we observe only 0.8\% illegal non-torso joint angles as compared to 1.4\% for~\cite{Zhou_2017_ICCV}.

The lower-half of Table~\ref{tab:symmetryTab} tabulates the standard deviation of bone lengths in mm across frames for SAP-Net and TP-Net. We can observe that TP-Net reduces the standard deviation of bone-length across the frames by 28.7\%. It is also worth noting that we do not use any additional filter (moving average, 1 Euro, etc.) which introduces lag and makes the motion look \textit{uncanny}. Finally, we present some qualitative results in Fig.~\ref{fig:coco_vis}, Fig.~\ref{fig:percentileAnalysis} and in the supplementary material to show that TP-Net effectively corrects the jerks in the poses predicted by SAP-Net.

\vspace{1.5em}

\begin{figure*}[t]
	\centering
	\includegraphics[width = \linewidth]{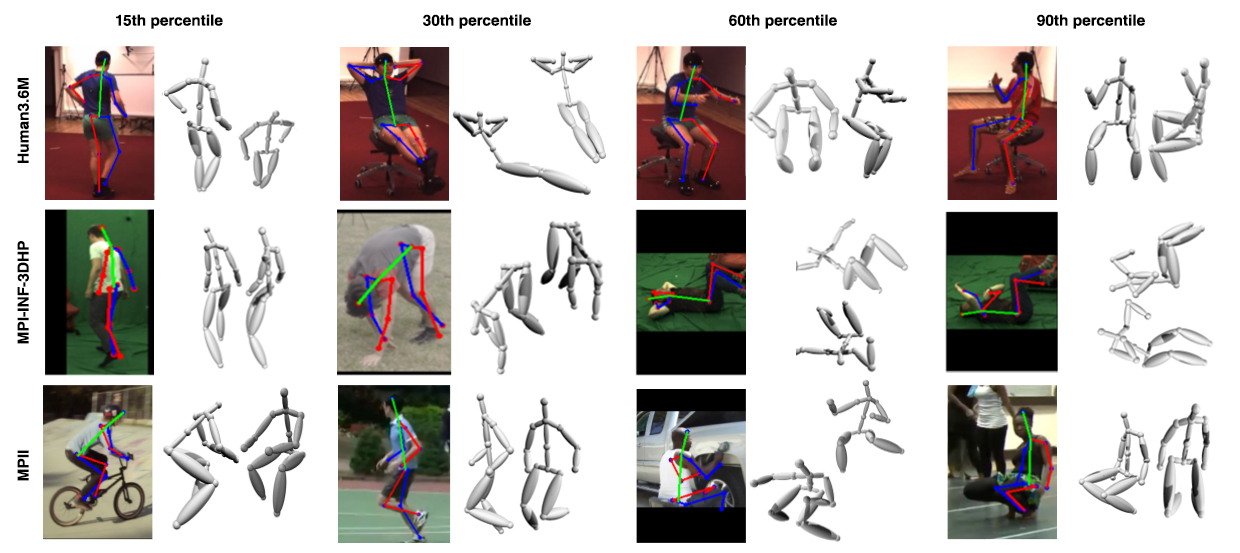}
	\vspace{-2em}
    \caption{Percentile analysis on Human3.6M (top row), MPI-INF-3DHP (middle row) and MPII (bottom row) datasets. The results are displayed at $15^{th}$, $30^{th}$, $60^{th}$ and $90^{th}$ percentile of error (MPJE for Human3.6M and MPI-INF-3DHP, 2D PCK for MPII) from left to right.}
    \label{fig:percentileAnalysis}
    \vspace{-2em}
\end{figure*}

\vspace{-02em}
\section{Conclusion}
\vspace{-1em}
We proposed two anatomically inspired loss functions, namely illegal-angle and symmetry loss. We showed them to be highly effective for training weakly-supervised ConvNet architectures for predicting valid 3D pose configurations from a single RGB image in-the-wild setting. We analyzed the evolution of local loss surfaces to clearly demonstrate the benefits of the proposed losses. We also proposed a simple, yet surprisingly effective, sliding-window fully-connected network for temporal pose modelling from a sequence of adjacent poses. We showed that it is capable of learning semantically meaningful short-term temporal and structure correlations. Temporal model was shown to significantly reduce jitters and noise from pose prediction for video sequences while taking $< 1ms$ per inference. Our complete pipeline improved the publised state-of-the-art by 11.8\% and 12\% on Human3.6M and MPI-INF-3DHP, respectively while running at 30fps on NVIDIA Titan 1080Ti GPU.
\vspace{2.5em}

{
\small
\bibliographystyle{ieee}
\bibliography{egbib}

\begin{thebibliography}{10}\itemsep=-1pt

\bibitem{akhter2015pose}
I.~Akhter and M.~J. Black.
\newblock Pose-conditioned joint angle limits for 3d human pose reconstruction.
\newblock In {\em CVPR}, 2015.

\bibitem{Alldieck2017}
T.~Alldieck, M.~Kassubeck, B.~Wandt, B.~Rosenhahn, and M.~Magnor.
\newblock Optical flow-based 3d human motion estimation from monocular video.
\newblock In {\em GCPR}, 2017.

\bibitem{andriluka14cvpr}
M.~Andriluka, L.~Pishchulin, P.~Gehler, and B.~Schiele.
\newblock 2d human pose estimation: New benchmark and state of the art
  analysis.
\newblock In {\em CVPR}, 2014.

\bibitem{bogo2016keep}
F.~Bogo, A.~Kanazawa, C.~Lassner, P.~Gehler, J.~Romero, and M.~J. Black.
\newblock Keep it smpl: Automatic estimation of 3d human pose and shape from a
  single image.
\newblock In {\em ECCV}, 2016.

\bibitem{casiez20121}
G.~Casiez, N.~Roussel, and D.~Vogel.
\newblock 1€ filter: a simple speed-based low-pass filter for noisy input in
  interactive systems.
\newblock In {\em SIGCHI}, 2012.

\bibitem{IonescuSminchisescu11}
C.~S. Catalin~Ionescu, Fuxin~Li.
\newblock Latent structured models for human pose estimation.
\newblock In {\em ICCV}, 2011.

\bibitem{Chen_2017_CVPR}
C.-H. Chen and D.~Ramanan.
\newblock 3d human pose estimation = 2d pose estimation + matching.
\newblock In {\em CVPR}, 2017.

\bibitem{ChenNie2013TIP}
J.~Chen, S.~Nie, and Q.~Ji.
\newblock Data-free prior model for upper body pose estimation and tracking.
\newblock {\em IEEE Transactions on Image Processing}, 22, 2013.

\bibitem{ChenWLSWTLCC16}
W.~Chen, H.~Wang, Y.~Li, H.~Su, Z.~Wang, C.~Tu, D.~Lischinski, D.~Cohen{-}Or,
  and B.~Chen.
\newblock Synthesizing training images for boosting human 3d pose estimation.
\newblock In {\em 3DV}, 2016.

\bibitem{cordts2016cityscapes}
M.~Cordts, M.~Omran, S.~Ramos, T.~Rehfeld, M.~Enzweiler, R.~Benenson,
  U.~Franke, S.~Roth, and B.~Schiele.
\newblock The cityscapes dataset for semantic urban scene understanding.
\newblock In {\em CVPR}, 2016.

\bibitem{Coskun_2017_ICCV}
H.~Coskun, F.~Achilles, R.~DiPietro, N.~Navab, and F.~Tombari.
\newblock Long short-term memory kalman filters: Recurrent neural estimators
  for pose regularization.
\newblock In {\em ICCV}, 2017.

\bibitem{he2016deep}
K.~He, X.~Zhang, S.~Ren, and J.~Sun.
\newblock Deep residual learning for image recognition.
\newblock In {\em CVPR}, 2016.

\bibitem{HERDA2005189}
L.~Herda, R.~Urtasun, and P.~Fua.
\newblock Hierarchical implicit surface joint limits for human body tracking.
\newblock {\em Computer Vision and Image Understanding}, 2005.

\bibitem{h36m_pami}
C.~Ionescu, D.~Papava, V.~Olaru, and C.~Sminchisescu.
\newblock Human3.6m: Large scale datasets and predictive methods for 3d human
  sensing in natural environments.
\newblock {\em IEEE TPAMI}, 2014.

\bibitem{Jahangiri:ICCV2017}
E.~Jahangiri and A.~L. Yuille.
\newblock Generating multiple diverse hypotheses for human 3d pose consistent
  with 2d joint detections.
\newblock In {\em ICCV}, 2017.

\bibitem{kingma2014adam}
D.~P. Kingma and J.~Ba.
\newblock Adam: A method for stochastic optimization.
\newblock In {\em ICLR}, 2015.

\bibitem{li20143d}
S.~Li and A.~B. Chan.
\newblock 3d human pose estimation from monocular images with deep
  convolutional neural network.
\newblock In {\em ACCV}, 2014.

\bibitem{Li_2015_ICCV}
S.~Li, W.~Zhang, and A.~B. Chan.
\newblock Maximum-margin structured learning with deep networks for 3d human
  pose estimation.
\newblock In {\em ICCV}, 2015.

\bibitem{Lin_2017_CVPR}
M.~Lin, L.~Lin, X.~Liang, K.~Wang, and H.~Cheng.
\newblock Recurrent 3d pose sequence machines.
\newblock In {\em CVPR}, 2017.

\bibitem{MSCOCO:2014}
T.~Lin, M.~Maire, S.~J. Belongie, L.~D. Bourdev, R.~B. Girshick, J.~Hays,
  P.~Perona, D.~Ramanan, P.~Doll{\'{a}}r, and C.~L. Zitnick.
\newblock Microsoft {COCO:} common objects in context.
\newblock {\em arXiv preprint arXiv:1405.0312}, 2014.

\bibitem{DBLP:journals/tog/LoperM0PB15}
M.~Loper, N.~Mahmood, J.~Romero, G.~Pons{-}Moll, and M.~J. Black.
\newblock {SMPL:} a skinned multi-person linear model.
\newblock {\em {ACM} Trans. Graph.}, 2015.

\bibitem{mono-3dhp2017}
D.~Mehta, H.~Rhodin, D.~Casas, P.~Fua, O.~Sotnychenko, W.~Xu, and C.~Theobalt.
\newblock Monocular 3d human pose estimation in the wild using improved cnn
  supervision.
\newblock In {\em 3DV}, 2017.

\bibitem{VNect_SIGGRAPH2017}
D.~Mehta, S.~Sridhar, O.~Sotnychenko, H.~Rhodin, M.~Shafiei, H.-P. Seidel,
  W.~Xu, D.~Casas, and C.~Theobalt.
\newblock Vnect: Real-time 3d human pose estimation with a single rgb camera.
\newblock In {\em ACM ToG}, 2017.

\bibitem{Moreno-Noguer_2017_CVPR}
F.~Moreno-Noguer.
\newblock 3d human pose estimation from a single image via distance matrix
  regression.
\newblock In {\em CVPR}, 2017.

\bibitem{NewellYD16}
A.~Newell, K.~Yang, and J.~Deng.
\newblock Stacked hourglass networks for human pose estimation.
\newblock In {\em ECCV}, 2016.

\bibitem{Park:2006}
M.~J. Park, M.~G. Choi, Y.~Shinagawa, and S.~Y. Shin.
\newblock Video-guided motion synthesis using example motions.
\newblock {\em ACM ToG}, 2006.

\bibitem{Pavlakos_2017_CVPR}
G.~Pavlakos, X.~Zhou, K.~G. Derpanis, and K.~Daniilidis.
\newblock Coarse-to-fine volumetric prediction for single-image 3d human pose.
\newblock In {\em CVPR}, 2017.

\bibitem{varunECCV2012}
V.~Ramakrishna, T.~Kanade, and Y.~Sheikh.
\newblock Reconstructing 3d human pose from 2d image landmarks.
\newblock In {\em ECCV}, 2012.

\bibitem{rogezNIPS}
G.~Rogez and C.~Schmid.
\newblock Mocap-guided data augmentation for 3d pose estimation in the wild.
\newblock In {\em NIPS}. 2016.

\bibitem{Rogez_2017_CVPR}
G.~Rogez, P.~Weinzaepfel, and C.~Schmid.
\newblock Lcr-net: Localization-classification-regression for human pose.
\newblock In {\em CVPR}, 2017.

\bibitem{2014arXiv1409.0575R}
O.~{Russakovsky}, J.~{Deng}, H.~{Su}, J.~{Krause}, S.~{Satheesh}, S.~{Ma},
  Z.~{Huang}, A.~{Karpathy}, A.~{Khosla}, M.~{Bernstein}, A.~C. {Berg}, and
  L.~{Fei-Fei}.
\newblock {ImageNet Large Scale Visual Recognition Challenge}.
\newblock {\em ArXiv e-prints}, 2014.

\bibitem{SARAFIANOS20161}
N.~Sarafianos, B.~Boteanu, B.~Ionescu, and I.~A. Kakadiaris.
\newblock 3d human pose estimation: A review of the literature and analysis of
  covariates.
\newblock {\em Computer Vision and Image Understanding}, 2016.

\bibitem{sminchisescu2003estimating}
C.~Sminchisescu and B.~Triggs.
\newblock Estimating articulated human motion with covariance scaled sampling.
\newblock {\em The International Journal of Robotics Research}, 2003.

\bibitem{Sun_2017_ICCV}
X.~Sun, J.~Shang, S.~Liang, and Y.~Wei.
\newblock Compositional human pose regression.
\newblock In {\em ICCV}, 2017.

\bibitem{Tome_2017_CVPR}
D.~Tome, C.~Russell, and L.~Agapito.
\newblock Lifting from the deep: Convolutional 3d pose estimation from a single
  image.
\newblock In {\em CVPR}, 2017.

\bibitem{urtasun2006temporal}
R.~Urtasun, D.~J. Fleet, and P.~Fua.
\newblock Temporal motion models for monocular and multiview 3d human body
  tracking.
\newblock {\em Computer vision and image understanding}, 2006.

\bibitem{Varol_2017_CVPR}
G.~Varol, J.~Romero, X.~Martin, N.~Mahmood, M.~J. Black, I.~Laptev, and
  C.~Schmid.
\newblock Learning from synthetic humans.
\newblock In {\em CVPR}, 2017.

\bibitem{Wei:2010}
X.~Wei and J.~Chai.
\newblock Videomocap: Modeling physically realistic human motion from monocular
  video sequences.
\newblock {\em ACM ToG}, 2010.

\bibitem{Nie_2017_ICCV}
B.~Xiaohan~Nie, P.~Wei, and S.-C. Zhu.
\newblock Monocular 3d human pose estimation by predicting depth on joints.
\newblock In {\em ICCV}, Oct 2017.

\bibitem{yasin2016dual}
H.~Yasin, U.~Iqbal, B.~Kruger, A.~Weber, and J.~Gall.
\newblock A dual-source approach for 3d pose estimation from a single image.
\newblock In {\em CVPR}, 2016.

\bibitem{Zhou_2017_ICCV}
X.~Zhou, Q.~Huang, X.~Sun, X.~Xue, and Y.~Wei.
\newblock Towards 3d human pose estimation in the wild: A weakly-supervised
  approach.
\newblock In {\em ICCV}, 2017.

\bibitem{zhou2016deep}
X.~Zhou, X.~Sun, W.~Zhang, S.~Liang, and Y.~Wei.
\newblock Deep kinematic pose regression.
\newblock In {\em ECCV Workshops}, 2016.

\bibitem{zhou2016sparseness}
X.~Zhou, M.~Zhu, K.~Derpanis, and K.~Daniilidis.
\newblock Sparseness meets deepness: 3d human pose estimation from monocular
  video.
\newblock In {\em CVPR}, 2016.

\bibitem{zhou2017brief}
Z.-H. Zhou.
\newblock A brief introduction to weakly supervised learning.
\newblock {\em National Science Review}, 2017.

\end{thebibliography}
}

\end{document}